\documentclass[10pt,twocolumn,letterpaper]{article}

\usepackage{cvpr}
\usepackage{times}
\usepackage{epsfig}
\usepackage{graphicx}
\usepackage{amsmath}
\usepackage{amssymb}
\usepackage{tabularx}

\usepackage[pagebackref=true,breaklinks=true,letterpaper=true,colorlinks,bookmarks=false]{hyperref}

\cvprfinalcopy 

\def\cvprPaperID{****} 

\ifcvprfinal\fi
\begin{document}

\title{Segmentation task for fashion and apparel}

\author{Hassler Castro\\
Universidad EAFIT\\
{\tt\small hcastro@eafit.edu.co}
\and
Mariana Ramirez\\
Universidad EAFIT\\
{\tt\small marami21@eafit.edu.co}
}

\maketitle

\begin{abstract}
The Fashion Industry is a strong and important industry in the global economy. Globalization has brought fast fashion, quick shifting consumer shopping preferences,  more competition, and abundance in fashion shops and retailers, making it more difficult for professionals in the fashion industry to keep track of what fashion items people wear and how they combine them.  This paper solves this problem by implementing several Deep Learning Architectures using the iMaterialist dataset consisting of 45,000 images with 46 different clothing and apparel categories
\end{abstract}

\section{Introduction}
The Fashion Industry owns 4 percent of the world market share and have a market value of 385.7 billion dollars \cite{fashionunited}. It is, indeed, a truly important industry for the global economy. 

However, with globalization comes innovation and competition, and it has become more and more challenging to gain and retain costumers; rapid changes in trends, fast fashion, quick shifting consumer shopping preferences  and the increase of fashion shops and retailers has made truly difficult for fashion designers and professionals to keep track of what fashion items people wear and how they combine them. 

Clothing and apparel recognition can help significantly in this problem. The ability to capture information about the type of fashion items that people wear and their combination in an efficient and cheap way can improve the business intelligence and increase the efficiency of companies in the fashion industry.

In this paper we present various semantic segmentation models which are capable of identifying multiple fashion items in pre-defined categories \textit{( e.g shirt, sleeve, sweater, jacket)}.

We organize the rest of this paper as follows: we talk about some related work in  object detection and semantic and instance segmentation in Section 2, we follow by giving the motivation that we had to realize this paper in Section 3, we give the objectives that we aim to fulfill in this paper in Section 4, we describe the methodology applied in Section 5, we later explore and understand the dataset used in Section 6, then we discuss different models used in Section 7, next we illustrate the results of thsee models along with the comparison between them in Section 8. Finally, we provide conclusions in Section 9.



\section{Related work}
Semantic Instance Segmentation (SIS), can be defined as the combination of two computer vision tasks, object detection and semantic segmentation. In the last years, both tasks have been resolved using Convolutional Neural Networks (CNN).

For object detection, among other implementations, R-CNN \cite{Girshick_2014} was developed, adopting a region proposal method for producing different instance proposals which are then fed into a CNN for classification. Fast R-CNN \cite{Girshick_2015} uses the technique of R-CNN with the difference that the convolutional layers of the CNNs are shared, which reduces the computational time from 50 seconds per image at test time to 2 seconds per image,  making it 10 times faster to train. Faster R-CNN \cite{ren2015faster} extract the region proposals used by the detector using the shared convolutional features, leading to a faster speed in object detection systems which results in 0.2 seconds per image at test time \cite{Dai_2016}

In 2014 a Simultaneous Detection and Segmentation (SDS) \cite{Hariharan_2014} method was proposed to solve the semantic segmentation challenge, it was built on top of RCNN, however, since it depends on a large number of region proposals it takes high computational resources \cite{Hu2018}. J. Long et. al. \cite{7298965} proposed a Fully Connected Network (FCN) method that improved performance over 10\% compared to the previous results on PASCAL CCOV 202 dataset, it consists mainly in replacing the fully connected layers in CNN into convolutional layers. Another architecture is the Unet which consists in a contracting path to capture context and a symmetric expanding path that enables precise localization \cite{Ronneberger_2015}, it was developed for Biomedical use, nevertheless, it has become widely applied in other areas. The University of Cambridge developed SegNet, a Fully Convolutional Neural Network consisting in an encoding network, a corresponding decoder network followed by a pixel-wise classification layer \cite{Badrinarayanan_2017}

The State of the Art for Instance Segmentation is Mask R-CNN which extends Faster R-CNN by adding a branch for predicting an object mask at the same time with the existing branch for bounding box recognition. It outperforms all existing instance segmentation and bounding-box object detectors \cite{He_2017}

\section{Motivation}
Visual analysis of clothing is a topic that has been increasingly gaining attention in the last years. The ability to recognize apparel products and 
 associated attributes in images have a wide variety of applications in e-commerce, online advertising, internet search and
in fashion design processes.

Clothing recognition has been addressed in different studies, \cite{6116276},  \cite{4587481}, \cite{Chao:2009:FRF:1631040.1631047}, however, little  studies address applications of apparel recognition in the fashion industry. 

Information such as how fashion products are being combined and how people combine clothing attributes could help professionals in the fashion industry increase the efficiency and the revenues of the industry. For example, Fashion forecasters can use this information to identify fashion changes and directions, by designers to create visually attractive lines, and by marketing executives to recognize selling points \cite{jia2018deeplearningbased} 

\section{Objectives}
\subsection{General Objective}
Develop a Deep Learning Algorithm that accurately assign segmentation and attribute labels for fashion images.
\subsection{Specific Objectives}
\begin{itemize}
    \item Develop different CNN models
    \item Compare the performance of the developed models
    \item Document findings
\end{itemize}

\section{Method}
To conduct the clothing and apparel segmentation the following steps were followed:
\begin{enumerate}
    \item Data acquisition: since the problem was presented as a Kaggle Challenge sponsored by Google AI, Fashionpedia and Samasource, the data was available in the challenge.
    \item Data exploration and preprocessing: to further understand the data we looked into the details of the dataset, from visualizing random images with its respective masks to understanding the classes distribution.
    In addition, we did some preprocessing on the data to improve the model efficiency (e.g normalization).
    \item Model selection: we selected the models for semantic segmentation that comply with a low or moderate amount of computational resources, that were developed in the last 10 years, and that achieved a high accuracy in different challenges and domains.
    \item Models training: we trained on 5 different models, Unet, SegNet,
    Atrous Resnet50 , FCN Resnet50 and DenseNet for image segmentation (adding the decoder path)
    \item Analysis and conclusions: Analysis of the performance of the different models were made along with conclusions of all the realized work
    
\end{enumerate}

\section{Dataset}
The iMaterialist (Fashion) 2019 dataset (iMFD), which we work on in this project, is available on https://www.kaggle.com/c/imaterialist-fashion-2019-FGVC6/data. \\
This dataset contains  50,000 clothing images (40,00 with apparel instance segmentation and 10,000 with both segmentation and fine-grained attributes) in daily-life, celebrity events, and online shopping, which were labeled by both domain experts and crowd workers. 

\begin{figure}
  \includegraphics[width=\linewidth]{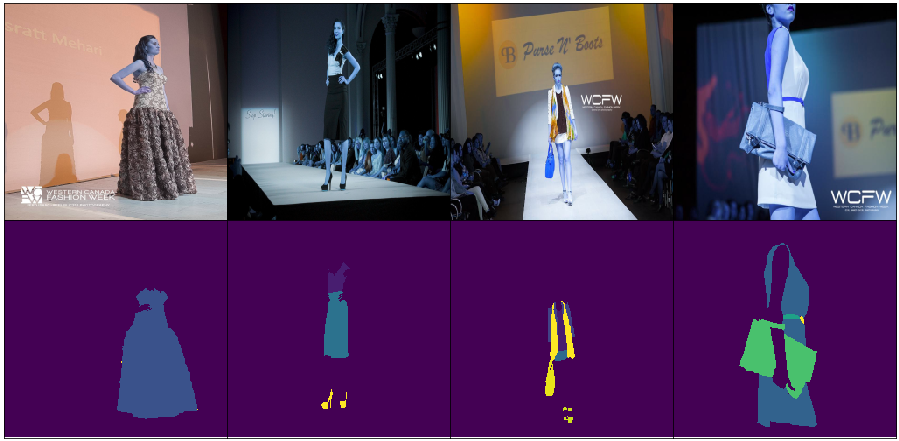}
  \caption{Desired output}
  \label{fig:masks}
\end{figure}

Figure \ref{fig:masks} shows the desired output on randomly chosen images

\begin{figure}
  \includegraphics[width=\linewidth, height=4cm]{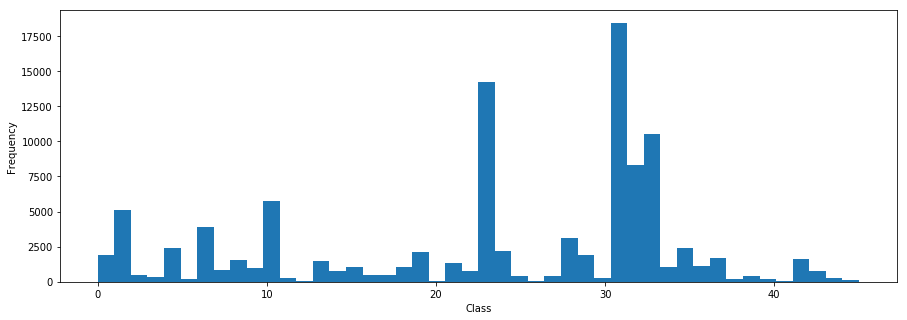}
  \caption{Histogram of iMaterialist dataset classes .}
  \label{fig:histogram_class}
\end{figure}

\subsection{Data exploration}
We proceed to further understand the iMaterialist dataset. Figure \ref{fig:histogram_class} illustrates the distribution of the classes, as it is shown it exist a class imbalance, class 31 (sleeve) appears 18,430 which is 17.8\% of all the images, while class 26 (umbrella) only appears 35 time showing in 0.034\% of the pictures.

\begin{figure}
  \includegraphics[width=\linewidth, height=4cm]{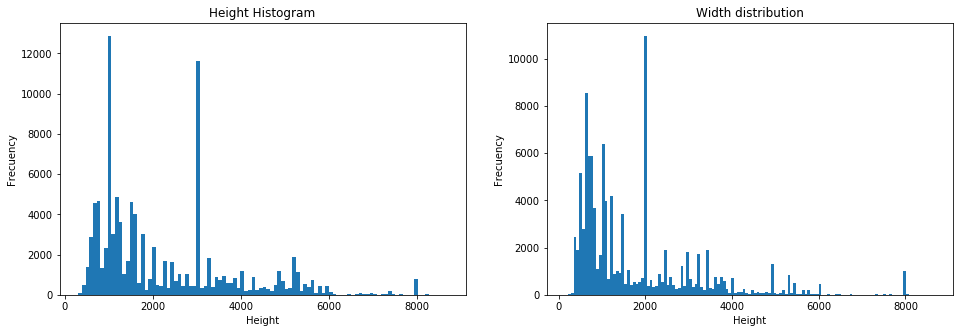}
  \caption{Histogram of iMaterialist dataset images weight and height respectively .}
  \label{fig:histogram_hw}
\end{figure}

In addition, we analyze the distribution of the weight and height of the images of the dataset as shown in Figure \ref{fig:histogram_hw}. As it can be seen, the weight and width of the images vary significantly from image to image, the biggest difference between the width and height of two images is of 5274 and 8384 pixels respectively

\subsection{Segmentation Masks}

For each image within the dataset, there is a set of descriptors responsible for defining 
which pixels within the image belong to a specific apparel, in this stage the segmentation masks are built
from these descriptors and flattened in a final matrix.

The descriptors are pairs of values ​​that contain a start position and a run length. E.g. '1 3' implies starting at pixel 1 and running to total of 3 pixels (1,2,3),
which means that when the masks are created for an image you get a matrix of shape (M x N x Z), where \textbf{M} is the width of the image, \textbf{N} the height of the image,
and \textbf{Z} the number of masks that the current image has (number of objects); Then, the masks are condensed in a matrix of shape (M x N), where Image$_{i, j}$ $\epsilon$ [0, 46] the number of classes.

Finally the images are standardized to a size of (256, 256) so their masks undergo the same transformation.

\section{Models}
\subsection{SegNet}
SegNet has an encoder network and a corresponding decoder \cite{Badrinarayanan_2017}
network, followed by a final pixel-wise classification layer, as shown in Table \ref{table:segnet} , the definition of each block can be found in Table
\ref{table:segnetblocks}.

\begin{table*}[!ht]
\begin{tabular}{|c|c|}
\hline
\textbf{Name}       & \textbf{Structure}                                                                                                                            \\ \hline
Simple\_Block       & (Conv -\textgreater reLU -\textgreater BatchNormalization) * 2 -\textgreater  MaxPooling                                                      \\ \hline
Complex\_Block      & (Conv -\textgreater reLU -\textgreater  BatchNormalization) * 3 -\textgreater  MaxPooling                                                     \\ \hline
Final\_Block        & (Conv -\textgreater reLU -\textgreater  BatchNormalization) * 2 -\textgreater  Conv(Num\_Classes) -\textgreater Reshape -\textgreater SoftMax \\ \hline
Simple\_Block\_Dec  & (Conv -\textgreater reLU -\textgreater BatchNormalization) * 2 -\textgreater  UpSampling                                                      \\ \hline
Complex\_Block\_Dec & (Conv -\textgreater reLU -\textgreater BatchNormalization) * 2 -\textgreater  UpSampling                                                      \\ \hline
\end{tabular}
\caption{SegNet Blocks}
\label{table:segnetblocks}
\end{table*}

\begin{table*}[ht]
\centering
\begin{tabular}{|c|c|c|c|c|c|c|}
\hline
\multicolumn{7}{|c|}{\textbf{SegNet}}                       \\ \hline
Type                   & Stride & Output Size & Kernel/FMap & Pool\_Size & Padding & \#Param \\ \hline
Simple\_Block\_1       & 1x1    & 128x128x64  & 3x3 / 64    & 2x2        & 'same'  & 39232   \\ \hline
Simple\_Block\_2       & 1x1    & 64x64x128   & 3x3 / 128   & 2x2        & 'same'  & 222464  \\ \hline
Complex\_block\_1      & 1x1    & 32x32x256   & 3x3 / 256   & 2x2        & 'same'  & 1478400 \\ \hline
Complex\_block\_2      & 1x1    & 16x16x512   & 3x3 / 512   & 2x2        & 'same'  & 5905920 \\ \hline
Complex\_block\_3      & 1x1    & 8x8x512     & 3x3 / 512   & 2x2        & 'same'  & 5905920 \\ \hline
UpSampling             & 2x2    & 16x16x512   & -           & -          & -       & 0       \\ \hline
Complex\_block\_Dec\_1 & 1x1    & 32x32x512   & 3x3 / 512   & 2x2        & 'same'  & 5905920 \\ \hline
Complex\_block\_Dec\_2 & 1x1    & 64x64x512   & 3x3 / 512   & 2x2        & 'same'  & 5905920 \\ \hline
Complex\_block\_Dec\_3 & 1x1    & 128x128x256 & 3x3 / 256   & 2x2        & 'same'  & 1478400 \\ \hline
Simple\_block\_Dec\_1  & 1x1    & 256x256x128 & 3x3 / 128   & 2x2        & 'same'  & 222464  \\ \hline
Final\_Block(softmax)  & 1x1    & 65536x47    & 3x3  / 64   & 2x2        & 'same'  & 114287  \\ \hline
\end{tabular}
\caption{SegNet Implementation}
\label{table:segnet}
\end{table*}

We believe that the inability of the model to increase its performance is due to the dataset inbalance, besides that it has several layers of max-pooling and sub-sampling that allow it to achieve more translation invariance for robust classification, correspondingly, there is a loss of spatial resolution of the feature maps. The increasingly lossy (boundary detail) image representation is not beneficial for segmentation where boundary delineation is vital \cite{Badrinarayanan_2017}. In our case this is reflected in lost of information of tiny objects (necklaces, neckline, sleeve and pockets) 


\subsection{U-Net}
It consists of a contracting path (left side) and an expansive path (right side). The contracting path follows the typical architecture of a convolutional network. It consists of the repeated application of two 3x3 convolutions (unpadded convolutions), each followed by a rectified linear unit (ReLU) and a 2x2 max pooling operation with stride 2 for downsampling. At each downsampling step we double the number of feature
channels. Every step in the expansive path consists of an upsampling of the feature map followed by a 2x2 convolution (“up-convolution”) that halves the number of feature channels, a concatenation with the correspondingly cropped feature map from the contracting path, and two 3x3 convolutions, each followed by a ReLU, these skip concatenation provide local information to global information while upsampling.

\subsection{ResNet50}
ResNet-50 is composed of 50 residual blocks, each of whom has a shortcut connection that skips one layer and performs identity mapping, which outputs are then added to the outputs of the stacked layers. In that way, all of the layers become learning residual functions with reference to the layer inputs. All of this is done to overcome the problem of vanishing/exploding gradients in Deep Convolutional Neural Networks (CNN) \cite{He_2016}

\subsection{Atrous ResNet50}
Atrous convolutions, enables to enlarge the field of views of filters to incorporate larger context. It also offers a way of controlling the  field of view and finds the best trade-off between accurate localization and context assimilation. \cite{chen2017rethinking}

Atrous ResNet50 is based on the same architecture of Resnet50 with the difference that it uses atrous convolutions to capture a long range of information in deeper blocks in an easier way. \cite{chen2017rethinking}

On Atrous ResNet50 we used transfer learning of weights that have been trained on segmentation problems in the PASCAL VOC dataset.

\subsection{FCN ResNet50}
Fully Convolutional Networks (FCN) are Convolutional Neural Networks (CNN) that take an input of some arbitrary size and produce correspondingly-sized output. This is achieved by replacing the last few layers of ordinary CNN with fully convolutional layers to make an efficient end-to-end learning and inference. 

FCN ResNet50 uses the same architecture of ResNet50 with the difference that the final classifier layer is removed and all fully connected layers are replaced by convolutions. \cite{Shelhamer_2017}

\subsection{DenseNet}
Dense Convolutional Network (DenseNet) is inspired in the observation that CNN can be deeper, more accurate, and efficient to train if they contain shorter connections between layers close to the input and close to the output. 

It achieves this by connecting all layers, with matching feature-map sizes directly with each other. In order to preserve the feed-forward fashion, each layer obtains additional inputs from all preceding layers and passes on its own feature-maps to all subsequent layers.

Densenet achieves better parameter efficiency as there is no need to relearn redundant feature maps. In addition, it improves the flow of information and gradients throughout the network, making it easier to train. \cite{Huang_2017}

\section{Results}

\subsection{Loss functions}

As mentioned above, class imbalance represents a big problem, and in segmentation tasks it is even harder to solve since the proportion of pixels belonging to the background is much higher than the proportion of pixels that belong to a non-background class.

To address this problem the first thing we came up with was to change the cost function so that the pixels that belong to any non-background class are more important than background class pixels, however, we had to be very careful with established weights because a poor initialization of weights could cause the network to take a long time to converge or simply stay stagnant in a state where the output images do not make any sense.

\textbf{Weighted softmax cross entropy with logits}

The two-class form of WCE can be expressed as

\begin {equation*}
    WCE = - \frac{1}{N} \sum_{n = 1} ^ {N} wr_ {n} log (p_n) + (1-r_n) log (1-p_n)
\end {equation*}

Where w is the weight vector to prioritize some classes over other
\begin{equation*}
    w = \frac {N- \sum_ {n} p_n} {\sum_ {n} p_n}
\end{equation*}

The weighted cross-entropy can be trivially extended to more than two classes. \cite{DBLP:journals/corr/SudreLVOC17}

This cost function expects the output of the network to be logits, i.e, a tensor that does not correspond to a probability distribution, so this tensor is then passed through an activation function, in this case softmax and finally the weights mentioned before are applied. 

We established previously, that weights are selected by the proportion of a class in the images, this means,in a set of images we get the average of class \textit{x} respect to the total number of pixel per image.
This way we give more importance to categories that appear less, and to pixels that belong to a less representative category within the same image.

\textbf{Generalized Dice Loss}

Generalized Dice Score (GDS) is a way of evaluating multiple class segmentation with a single score but it has not been used in the context of discriminative model training. It takes the form:

\begin {equation*}
    GDL = 1-2 \frac{\sum_{l = 1}^{2} w_l \sum_{n} r_{ln} p_{ln}} {\ sum_ {l = 1}^{2} w_l \sum_{ n} r_{ln} + p_{ln}}
\end{equation*}

where w$_{l}$ is used to provide invariance to different label set properties. In the following, we adopted the notation GDL${_v}$ when
\begin {equation*}
    w_l = \frac {1} {(\sum_{n = 1} ^ {N} r_{ln}) ^ 2}
\end {equation*}

When choosing the GDL weighting, the contribution of each label is corrected by the inverse of its volume, thus reducing the well-known correlation between region size and Dice score. \cite{DBLP:journals/corr/abs-1708-02002}

This function allowed us to deal with the unbalanced class problem in a less robust way than the previously mentioned function
Dice loss performed relatively well on the iMaterialist Dataset, however there was no significant improvement with respect to the results obtained with \textit{weighted cross entropy loss} and it can be noticed in Dice derivative, that the gradient applied in the GDL is much more complex than the gradient of \textit{weighted softmax cross entropy}.

GDL gradient:

\begin {equation*}
\begin{split}
& \frac{\partial GDL}{\partial p_i} = - 2 \frac{(w_1 ^ 2 - w_2 ^ 2)[\sum_ {n = 1}^{N} p_nr_n-r_i \sum_{n = 1}^{N}(p_n + r_n)]} {[(w_1-w_2) \sum_{n = 1}^{N} (p_n + r_n) + 2Nw_2] ^ 2 } \\
& + \frac{Nw_2 (w_1 + w_2) (1-2r_i)} {[(w_1-w_2) \sum_{n = 1}^{N} (p_n + r_n) + 2Nw_2] ^ 2 }
\end{split}
\end {equation*}

\textbf{Focal loss}
Focal loss propose  reshaping the loss function to down-weight easy examples and thus focus training on hard negatives. More formally, proposes to add to modulating factor;

\begin {equation*}
    (1-p_t) ^ {\ gamma}
\end {equation*}

to the cross entropy loss, with tunable focusing parameter
\begin {equation*}
    \gamma \geq 0
\end {equation*}
We define the focal loss as:

\begin{equation*}
    FL (p_t) = - (1-p_t) ^ {\ gamma} log (p_t)
\end{equation*}

Focal loss is a robust loss function that aims to give more importance to classes that appear less and therefore makes a good choice for our problem.

\subsection{Metrics}

\textbf{Accuracy}

This is the metric that in many cases is used as  default, however, for the problem we are dealing with (\textit{multiclass segmentation}) where the background in the images occupies a significant proportion, this metric is not enough, it is a blind metric, because a model that only predicts \textit{background} will have a fairly high accuracy, and the results obtained clearly are wrong.

\textbf{IoU}

Intersection over union is related to the aforementioned coefficient of dice that measures the number of pixels that are in common between the target mask and the predicted mask, this value is divided by the total number of pixels present in both masks. 
The way we approached IoU was by using binary IoU, which means that we only had two classes, non-background and background, in this way, the model will get a high IoU if it succeed differentiating any non-background class from the background, this means that it will have a high IoU even if it fails at predicting the ground truth non-background class

IoU goes together with accuracy because it is not enough to have a high IOU without a high accuracy, that is, not only establishing that there is an object but it must be precise when classifying which object it is.

\subsection{Optimizer}

\textbf{Adam optimizer}
Adam optimizer is a method for efficient stochastic optimization that only requires first-order gradients with little memory requirement. The method computes individual adaptive learning rates for different parameters from estimates of first and second moments of the gradients\cite{2014arXiv1412.6980K}

ADAM optimizer is considered the state of the art when it comes to convolutions or images, because through different domains it has shown very good results. It was also the first optimizer to be tested and we blindly trust his performance. However Table \ref{table:adamptimizer} shows poor results, we believe this low performance (IoU) is due to the fact that it takes a step on the loss function with the update average gradients of many examples, going into the direction of the dominant class (background). 

\begin{table}
\centering
\caption{Metrics, using AdamOptimizer}
\label{table:adamptimizer}
\begin{tabular}{|c|c|c|c|l|} 
\hline
\multicolumn{5}{|c|}{\textbf{Metrics (AdamOptimizer) }}                                                                                                                                                                                                                                            \\ 
\hline
Model                                                                        & \begin{tabular}[c]{@{}c@{}}LR = 0.01 \\Accuracy\end{tabular} & \begin{tabular}[c]{@{}c@{}}LR=0.001 \\Accuracy\end{tabular} & \begin{tabular}[c]{@{}c@{}}LR=0.01 with\\ Decay = 1e-6 \\Accuracy\end{tabular} & IoU   \\ 
\hline
U-Net                                                                        & 80.84                                                        & 76.75                                                       & 76\%                                                                           & 0.0   \\ 
\hline
SegNet                                                                       & 78.02                                                        & 77.8                                                        & 77\%                                                                           & 0.06  \\ 
\hline
\multicolumn{1}{|l|}{\begin{tabular}[c]{@{}l@{}}Atrous\\ResNet\end{tabular}} & \multicolumn{1}{l|}{~ ~ 79.7}                                & \multicolumn{1}{l|}{~ ~ ~77.4}                              & \multicolumn{1}{l|}{~ ~ ~ ~ 77\%}                                              & 0.09  \\
\hline
\end{tabular}
\end{table}

\textbf{SGD (Stochastic gradient descent)}
It was by far the best optimizer that we tried. SGD worked better than Adam because to take a step inside the cost function, you only need to see one example at a time and not all the batch size which allows to go in a direction that is not strongly dominated by the background pixels unlike Adam or RMSprop optimizers.

\begin{figure}
  \includegraphics[width=\linewidth]{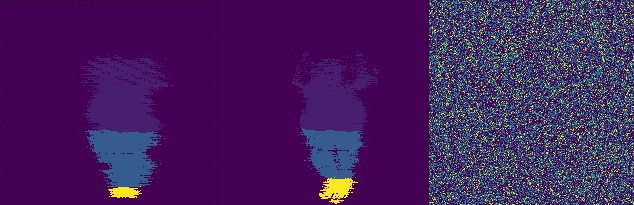}
  \caption{ \textbf{(a)} Shows a predicted mask using weighted categorical cross entropy, \textbf{(b)} Focal loss function and \textbf{(c)} result of making a bad weight estimation }
  \label{fig:results}
\end{figure}

\subsection{Method}

With the previously tested metrics that define how well does a model perform, as well as with the cost functions described above and choosing SGD as optimizer, since this outperformed Adam and RMSprop, we can introduce the training method that guided the development
of the project.

\textbf{1. Overfit a complex example}

Its objective is to verify which models are capable of completing the task, and which are better completing it, the models trained with 300 epochs on the same image in different models,  with different hyperparameters, the best results can be seen in Figure \ref{fig:comparison_overfitted}

\begin{figure}
  \includegraphics[width=\linewidth]{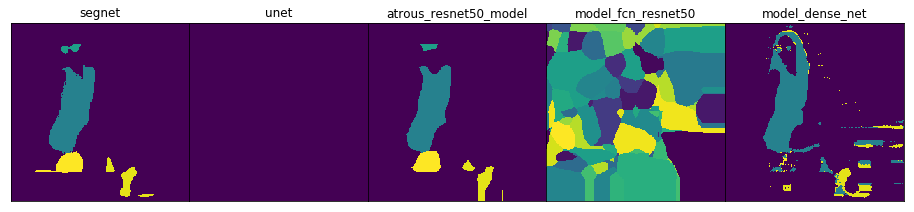}
  \caption{Model comparison}
  \label{fig:comparison_overfitted}
\end{figure}

The models that did not give decent results are discarded, however, it is  future work to inspect why, however, it is beyond the scope of this paper.

\textbf{2. Overfit a subset with high variety}

A subset where all classes can be found in similar proportions to the complete dataset, i.e unbalanced.
This step allowed choosing \textbf{quickly} which optimizer and cost functions were better for each model by validating with another training subset with same characteristics Table \ref{table:coptselection}

\begin{table}[h]
\begin{tabular}{|c|c|c|}
\hline
Model           & Optimizer & Cost Function  \\ \hline
SegNet          & SGD       & Focal Loss     \\ \hline
Atrous ResNet50 & SGD       & W Crossentropy \\ \hline
DenseNet & SGD       & W Crossentropy \\ \hline
\end{tabular}
\caption{Cost function and optimizer selection}
\label{table:coptselection}
\end{table}

\textbf{3. Tuning hyperparameters}

With the whole training and validation set, we considered that each pixel counted as an example so we split the dataset into 90\% training and 10\% validation.
We started tuning hyperparameters to obtain better results on each model, SegNet and Atrous ResNet50 were better than DenseNet (Higher accuracy and IoU while training) so we only spent time tuning these two models.
Figure \ref{fig:results} shows some predicted masks.

\begin{figure}
  \includegraphics[width=\linewidth]{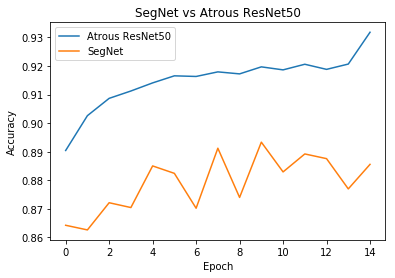}
  \caption{Accuracy comparison}
  \label{fig:accVs}
\end{figure}

\begin{figure}
  \includegraphics[width=\linewidth]{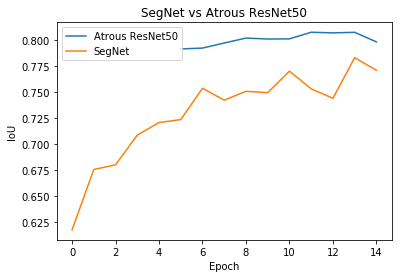}
  \caption{IoU comparison}
  \label{fig:IOUvs}
\end{figure}

After training both models for days, we got the following results, on new data. Figure \ref{fig:accVs} \ref{fig:IOUvs} Atrous ResNet did a better generalizing on new examples than SegNet, even when SegNet got better results on the training and validation datasets. Table \ref{table: iou_acc} shows the IoU and the accuracy of both models during testing. We think this Atrous ResNet robustness is due to the fact that it started to train with pretrained weights.
Experiments were made using a Nvidia Tesla K80 GPU.

\begin{table}[h]
\begin{tabular}{|c|c|c|}
\hline
Model            & Accuracy & IoU  \\ \hline
Atrous ResNet-50 & 0.93     & 0.8  \\ \hline
SegNet           & 0.89     & 0.77 \\ \hline
\end{tabular}
\caption{Accuracy and IoU on testing in Atrous ResNet-50 and SegNet}
\label{table: iou_acc}
\end{table}

\begin{figure}
  \includegraphics[width=\linewidth]{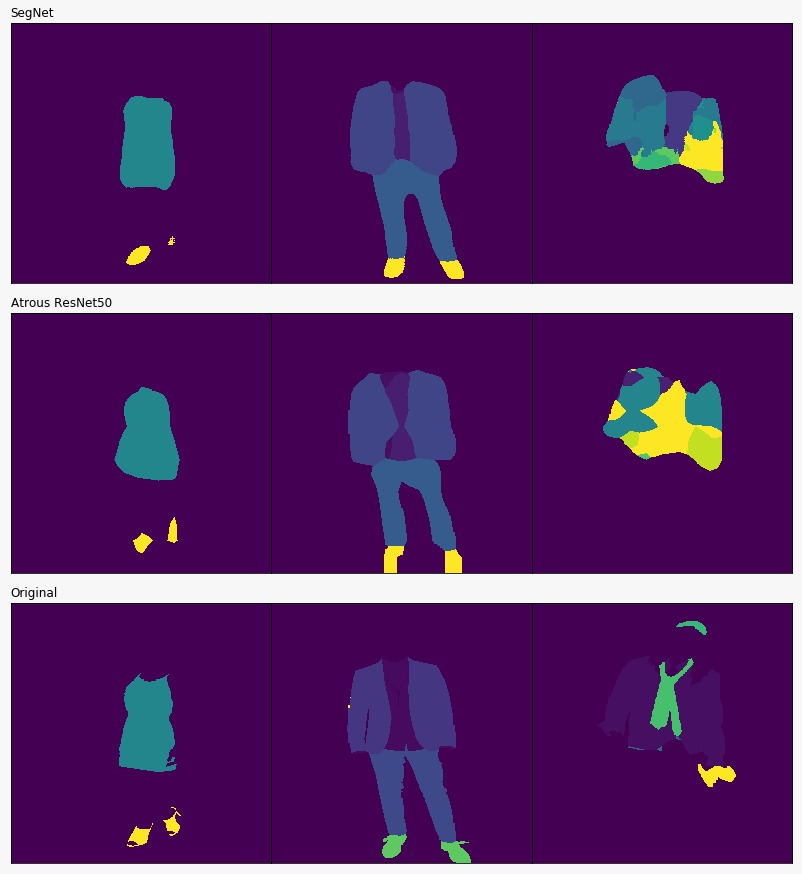}
  \caption{Final results}
  \label{fig:comparison_overfitted}
\end{figure}

\section{Conclusions}

The development of the project led us to important conclusions that will prevent us from commenting on errors in the future in deep learning projects, in addition we learned that intuition is slower than trial and error.
\textbf{1.} Fast iteration is essential, establish a set of metric cost functions, optimizers and models and quickly group the possible combinations to overfit an example and a group of examples that represent the dataset. This will prevent your intuition from blinding you of trying new things and will save important time.
\textbf{2.} Transfer learning, can become a great ally to address problems with class imbalance, and make your models more robust i.e better generalization capacity.
\textbf{3.} Even a simple model can get decent results with the right parameter selection.
\textbf{4.} Deep Segmentation models tend to do well in classifying large objects, however, its accuracy diminish as the objects to be detected get smaller, we believe this is because of the loss of spatial information that occur during the downsampling 
\textbf{5.} Class imbalance can affect dramatically the performance of the model, one of many ways one can solve this is by changing the loss function to give more weight to the underrepresented classes 
\textbf{6.} Accuracy is a widely used metric, nevertheless, it does not guarantee that the model is predicting well and it should not be used by itself when the dataset suffers from class imbalance. 


{\small
\bibliographystyle{ieee}
\bibliography{egbib}
}

\end{document}